\def\year{2017}\relax
\documentclass[letterpaper]{article} %DO NOT CHANGE THIS
\usepackage{aaai17}  %Required
\usepackage{times}  %Required
\usepackage{helvet}  %Required
\usepackage{courier}  %Required
\usepackage{url}  %Required
\usepackage{graphicx}  %Required
\usepackage{hyperref}
\begin{document}
\title{MengeROS: a Crowd Simulation Tool for Autonomous Robot Navigation}
\author{Anoop Aroor\textsuperscript{1}, Susan L. Epstein\textsuperscript{1,2} and Raj Korpan\textsuperscript{1}\\
Department of Computer Science\\
The Graduate Center\textsuperscript{1} and Hunter College\textsuperscript{2} of The City University of New York\\
New York, NY 10065\\
aaroor@gradcenter.cuny.edu, susan.epstein@hunter.cuny.edu, rkorpan@gradcenter.cuny.edu
}
\maketitle
\begin{abstract}
While effective navigation in large, crowded environments is essential for an autonomous robot, preliminary testing of algorithms to support it requires simulation across a broad range of crowd scenarios. Most available simulation tools provide either realistic crowds without robots or realistic robots without realistic crowds. This paper introduces MengeROS, a 2-D simulator that realistically integrates multiple robots and crowds. MengeROS provides a broad range of settings in which to test the capabilities and performance of navigation algorithms designed for large crowded environments.
\end{abstract}

%scenario=plan  production,init, strategy,map

\noindent 
Robots are increasingly deployed in crowded indoor environments, such as museums, shopping malls, and conference centers \cite{Tsui:2011:EUC:1957656.1957664}. Before such a deployment, the robot's navigation control must be tested extensively, particularly to prevent collision. It is challenging, however, to design and execute appropriate, large-scale, real-world testing for robots across the broad range of crowd conditions that arise in service areas. A shopping-mall crowd, for example, varies with the time, the day of the week, and irregularly scheduled events, and each permutation defines a different test. A flexible, accurate crowd simulator is thus essential before a robot is deployed in a crowded environment. Such a simulator can also support the evaluation of different navigation algorithms under comparable crowd conditions. This paper introduces a novel tool, \emph{MengeROS}, that integrates a flexible, open-source crowd simulator called \emph{Menge} \cite{curtis2016menge} with \emph{ROS}, the standard operating system for robots that navigate.

Specialized robot simulators (e.g., Gazebo \cite{koenig2004design} or Stage \cite{gerkey2003player}) do not simulate realistic crowds. Moreover, most crowd simulators (e.g., PedSim\footnote{\url{http://pedsim.silmaril.org}}, OpenSteer \cite{reynolds1999steering}, Menge, and Continuum \cite{treuille2006continuum}) do not simulate robots. Only two known crowd simulators include robots: PedSim\_ros\footnote{\url{https://github.com/srl-freiburg/pedsim_ros}}, which is restricted to one collision avoidance model, and a ROS version of Continuum, which is not freely available for research. 

In contrast, MengeROS is a flexible, open-source 2-D crowd simulator. MengeROS is not meant to simulate sophisticated full-body human simulation such as gestures and expressions, it is designed to facilitate research in multi-robot path planning problems in large crowded environments. To the best of our knowledge, MengeROS is the only open-source simulator that supports the movement of multiple robots through a broad range of crowd scenarios.

\section{The Menge Crowd Simulator}
Menge's \textit{crowd scenario} includes a map of the environment's static elements (e.g., walls, furniture), the number of simulated people (\emph{pedestrians}) in the crowd, their initial locations, along with one decision-making strategy and one collision-avoidance strategy applied to all pedestrians. Menge describes how each pedestrian determines its goals, and how all of them select their next moves and avoid collisions with one another. A \emph{location} in Menge is either a pair of coordinates (\emph{x,y}) or a delineated area in the environment (e.g., the kitchen). \emph{Goal selection} specifies a \emph{target sequence} (locations to visit) for each pedestrian. \emph{Plan computation} specifies how Menge calculates the next location for all pedestrians as they move toward their targets. Both implemented methods, A* and potential fields, generate and assign a \emph{velocity vector} (direction and distance) to each pedestrian. A* pursues an optimal shortest path; potential fields uses an attractor mechanism. \emph{Collision avoidance} adjusts each pedestrian's intended velocity vector to prevent collisions with the other pedestrians.

Menge currently has six collision avoidance models, out of which four are based on \emph{social force model} \cite{helbing1995social}, where nearby objects or pedestrians attract or repel a pedestrian, whose revised velocity vector reflects the result of all those forces. Remaining two models, ORCA and PedVO are based on velocity obstacles. The velocity obstacle (\emph{VO}) of a pedestrian is the set of all velocity vectors that will result in collision. Collision-free motion requires that every pedestrian have a velocity vector outside its VO. To prevent livelock and find an optimal solution, ORCA shares this responsibility equally among all pedestrians. PedVO adapts ORCA to behave more similarly to people. 

In Menge, users can select among precoded options for goal selection, plan computation, and collision avoidance, or implement their own. As a result, Menge can simulate many different kinds of crowd scenarios. This flexibility is a significant improvement over earlier crowd simulators, which hardcoded a single approach. Nonetheless, Menge is not available through ROS and does not simulate robots.

\section{MengeROS}
A typical ROS-based robot navigation framework uses a \emph{simulator node}. This node accepts as input a velocity command in a ROS-specified format, and returns simulated sensor readings (e.g., laser range scans) at a specified frequency. To determine the robot's motion, a controller node generates velocity commands based on the most recent sensor reading it has received from the simulator node. 

MengeROS simulates both robots and pedestrians in a single node. MengeROS controls pedestrian behavior just as Menge does. It also allows multiple robots to be introduced, each with its own external controller. A robot in MengeROS executes the velocity commands received from its external ROS controller. This is similar to the way other robot simulators (e.g., Gazebo or Stage) interface with ROS. Pedestrians avoid the robot and one another with the collision avoidance option specified in the Menge control files. A robot, however, is completely dependent on the external commands from its own controller for collision avoidance.

\begin{figure}
\includegraphics[scale=0.211]{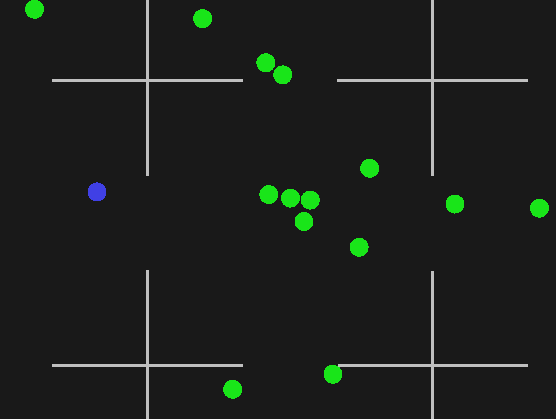}
\includegraphics[scale=0.314]{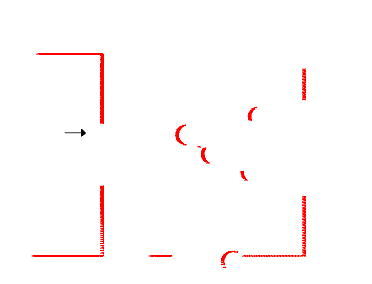}
\includegraphics[scale=0.7]{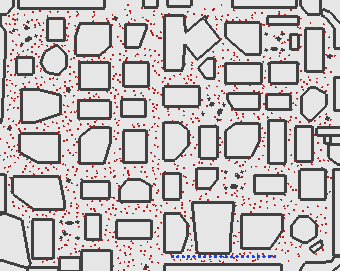}
\caption{(top left) Aerial view of a simple world with 1 robot and 14 pedestrians. (top right) Robot range scanner readings. (bottom) A trade show world with 1000 pedestrians and 20 robots in a row at the lower right.}
\end{figure}

MengeROS can simulate a laser scanner mounted on a robot. The top of Figure 1 shows two aerial views of a simple world. On the left is the ground truth, with 1 blue robot and 14 green pedestrians. On the right is the robot's view when it is located at the arrow's tail and oriented toward its head. Distances to obstacles are reported by a simulated laser with a (configurable) 220$^{\circ}$ field of view with maximum (configurable) range of 25 meters. MengeROS returns the positions of all pedestrians and robots in ROS-compatible format, for use by all other ROS nodes. 

\section{Results and Discussion}
MengeROS readily simulates large crowds, including 1000 pedestrians that move simultaneously in Menge's complex trade show environment, shown at the bottom of Figure 1. Each decision cycle computes and assigns a new velocity vector to every pedestrian. On an 8-core, 1.2 GHz workstation, 100 decision cycles without robots average 51ms each. Because each robot's range sensors must be processed separately, robots slow performance. This slowdown appears to be linear in the number of robots. Average decision cycle times with 5, 10, 15, and 20 robots were 437ms, 824ms, 1204ms and 1568ms, respectively.

Currently, MengeROS assumes that all robots are circular (with a configurable radius) and that the laser scan data and the robot's actions are noise-free. Future work will introduce noise, simulate robots of different shapes, process data from sensors other than range finders, reduce decision cycle time, and introduce personalized human reactions to robots. Code for MengeROS along with documentation, examples, and demos are hosted on GitHub\footnote{\url{https://github.com/ml-lab-cuny/menge_ros/}}. 

Recent work to improve navigation in a crowded environment utilized MengeROS to simulate robot movement through up to 90 pedestrians \cite{anoop2017crowds}. In this work, a robot learned a \emph{crowd density map} of the moving pedestrians from the simulated 2-D laser scan data. Multiple simulations with MengeROS also allowed for easy comparison of traditional A* with the proposed method, CSA*, which uses crowd density maps to improve navigation performance. 

In summary, robots in crowded indoor environments experience new challenges as their navigation algorithms confront dynamic obstacles. Research costs to develop algorithms in realistic scenarios can be significantly reduced by simulation. MengeROS is an efficient, flexible and extensible new tool for such work. It builds on the Menge crowd simulator, and allows robotics researchers to test their algorithms in realistic crowds before deployment. 

%\section{No space}
%Menge offers six different models for plan adaptation: social forces \cite{helbing1995social}, PedVO \cite{curtis2014pedestrian}, ORCA \cite{vandenBerg2011}, Johanesson et al\cite{johansson2007specification}, Karamouza et al \cite{karamouzas2009predictive}, and Zanlungo \cite{zanlungo2011social}

\bibliographystyle{aaai}
\bibliography{references}

\begin{thebibliography}{}

\bibitem[\protect\citeauthoryear{Aroor and Epstein}{2017}]{anoop2017crowds}
Aroor, A., and Epstein, S.~L.
\newblock 2017.
\newblock {Towards Crowd Sensitive Path Planning}.
\newblock In {\em AAAI 2017 Fall symposium on Human-Agent Groups}.

\bibitem[\protect\citeauthoryear{Curtis, Best, and
  Manocha}{2016}]{curtis2016menge}
Curtis, S.; Best, A.; and Manocha, D.
\newblock 2016.
\newblock {Menge: A Modular Framework For Simulating Crowd Movement}.
\newblock {\em Collective Dynamics} 1:1--40.

\bibitem[\protect\citeauthoryear{Gerkey, Vaughan, and
  Howard}{2003}]{gerkey2003player}
Gerkey, B.; Vaughan, R.~T.; and Howard, A.
\newblock 2003.
\newblock {The Player/Stage Project: Tools For Multi-Robot And Distributed
  Sensor Systems}.
\newblock In {\em Proceedings of the 11th international conference on advanced
  robotics}, volume~1,  317--323.

\bibitem[\protect\citeauthoryear{Helbing and Molnar}{1995}]{helbing1995social}
Helbing, D., and Molnar, P.
\newblock 1995.
\newblock {Social Force Model for Pedestrian Dynamics}.
\newblock {\em Physical review E} 51(5):4282.

\bibitem[\protect\citeauthoryear{Koenig and Howard}{2004}]{koenig2004design}
Koenig, N., and Howard, A.
\newblock 2004.
\newblock {Design and Use Paradigms For Gazebo, An Open-Source Multi-Robot
  Simulator}.
\newblock In {\em Proceedings of IEEE/RSJ International Conference on
  Intelligent Robots and Systems}, volume~3,  2149--2154.

\bibitem[\protect\citeauthoryear{Reynolds}{1999}]{reynolds1999steering}
Reynolds, C.~W.
\newblock 1999.
\newblock {Steering Behaviors For Autonomous Characters}.
\newblock In {\em Game developers conference}, volume 1999,  763--782.

\bibitem[\protect\citeauthoryear{Treuille, Cooper, and
  Popovi{\'c}}{2006}]{treuille2006continuum}
Treuille, A.; Cooper, S.; and Popovi{\'c}, Z.
\newblock 2006.
\newblock {Continuum Crowds}.
\newblock In {\em ACM Transactions on Graphics (TOG)}, volume~25,  1160--1168.

\bibitem[\protect\citeauthoryear{Tsui \bgroup et al\mbox.\egroup
  }{2011}]{Tsui:2011:EUC:1957656.1957664}
Tsui, K.~M.; Desai, M.; Yanco, H.~A.; and Uhlik, C.
\newblock 2011.
\newblock {Exploring Use Cases for Telepresence Robots}.
\newblock In {\em Proceedings of the 6th International Conference on
  Human-robot Interaction}, HRI '11,  11--18.

\end{thebibliography}

\end{document}